\documentclass{sigkddExp}
\pdfoutput=1

 \everycr={\noalign{\global\advance\stepno by 1}}%
\usepackage{amsmath}
\usepackage[breaklinks]{hyperref}
\usepackage{algorithm2e}

\usepackage{graphicx}
\usepackage{enumerate}
\usepackage{biblatex}
\usepackage{comment}
\usepackage{url}
\usepackage{xspace}
\usepackage{float}
\usepackage{enumitem}
\usepackage{tikz}
\usepackage{caption}
\usepackage{array}
\usepackage{fancyhdr}
\usepackage{booktabs}
\usepackage{multirow}
\usepackage{rotating}
\usepackage{longtable}
\usepackage{makecell}
\usepackage{color}
\usepackage{colortbl}
\usepackage{bigdelim}
\usepackage{amsmath}

\DeclareUnicodeCharacter{2212}{-}

\newlist{questions}{enumerate}{2}
\setlist[questions,1]{label=\textbf{RQ\arabic*.},ref=\textbf{RQ\arabic*}}
\setlist[questions,2]{label=(\alph*),ref=\thequestionsi(\alph*)}

\begin{document}
%

\title{Source-Free Domain Adaptation via Multi-view Contrastive Learning }

\numberofauthors{3}

\author{
\alignauthor Amirfarhad Farhadi\thanks{These authors contributed equally to this work.} \\
   \affaddr{School of Computer engineering, Iran university of Science and Technology, Tehran, Iran} \\
    \email{am\_farhadi@mail.iust.ac.ir}
\alignauthor Naser Mozayani \\
    \affaddr{School of Computer engineering, Iran university of Science and Technology, Tehran, Iran} \\
    \email{mozayani@iust.ac.ir}
    \\ 
 \alignauthor Azadeh Zamanifar \\
    \affaddr{Department of Computer Engineering, SR.C, Islamic Azad university} \\
    \email{azamanifar@srbiau.ac.ir} \\  
}
\maketitle
\begin{abstract}
Domain adaptation has become a widely adopted approach in machine learning due to the high costs associated with labeling data. It is typically applied when access to a labeled source domain is available. However, in real-world scenarios, privacy concerns often restrict access to sensitive information, such as fingerprints, bank account details, and facial images. A promising solution to this issue is Source-Free Unsupervised Domain Adaptation (SFUDA), which enables domain adaptation without requiring access to labeled target domain data. Recent research demonstrates that SFUDA can effectively address domain discrepancies; however, two key challenges remain: (1) the low quality of prototype samples, and (2) the incorrect assignment of pseudo-labels. To tackle these challenges, we propose a method consisting of three main phases. In the first phase, we introduce a Reliable Sample Memory (RSM) module to improve the quality of prototypes by selecting more representative samples. In the second phase, we employ a Multi-View Contrastive Learning (MVCL) approach to enhance pseudo-label quality by leveraging multiple data augmentations. In the final phase, we apply a noisy label filtering technique to further refine the pseudo-labels. Our experiments on three benchmark datasets—VisDA-2017, Office-Home, and Office-31—demonstrate that our method achieves approximately 2\% and 6\% improvements in classification accuracy over the second-best method and the average of 13 well-known state-of-the-art approaches, respectively.

\end{abstract}


\section{Introduction} \vspace{0.5cm}
Recently, transfer learning has become one of the most trending topics in the field of machine learning. Domain Adaptation (DA) is a branch of transfer learning, and in recent years, DA approaches have gained more popularity due to the cost of labeling and knowledge deficiency. Some recent  in the context of DA, when labeled data is available, it is referred to as the source domain, whereas when labeled data is not available, it is called the target domain \cite{farhadi2024leveraging, farhadi2024domain}. Generally, the main goal of DA algorithms is to adapt the source domain (with labeled data) to the target domain (with unlabeled data) when the distributions are different but relevant. In traditional DA assume full access to samples during training phase, however in real world typically not allow to use some information and bio-metric data because of privacy \cite{pan2025overcoming}. Currently, data security and the protection of personal information are crucial subjects \cite{farhadi2024leveraging1, taheri2024enhancing}. One of the challenging aspects in data analysis has been accessing sensitive data. In real world problems, frequently, we do not have access to the source domain and only have data from the target domain along with pre trained models \cite{li2024comprehensive}. Moreover, One of the primary challenges arising from the ever growing size of datasets is the associated computational cost and memory limitations \cite{wang2025robust, tang2024source}. To address these constraints while preserving privacy and efficiency, SHOT \cite{liang2020we} was the first to introduce the concept of Source-Free Unsupervised Domain adaptation(SFUDA) and employed a self-supervised pseudo-labeling strategy. It calculated target domain prototypes using a clustering algorithm and enabled the well trained source model to assign pseudo labels to target domain samples based on these prototypes. Some methods, such as \cite{ye2021source, liu2025adaptive}, attempt to convert SFUDA into a conventional Unsupervised Domain Adaptation(UDA) problem by estimating the source domain's data distribution through latent knowledge extracted from trained samples in the source domain. However, this estimated distribution may significantly differ from the true distribution, which could lead to a decrease in model accuracy. Another effective solution involves generating pseudo labels for the unlabeled target data in an iterative manner based on the well trained source model. This approach avoids the unreliability that may arise from directly predicting source domain data and has become a common strategy in SFUDA tasks \cite{liang2020we,liang2021source, qiu2021source, xu2025unraveling}. One of the first methods in SFUDA is \cite{kim2021domain} based on self-entropy in target data, with the impact of self entropy being a crucial part of the method, which we leverage in our work. \cite{ma2024source} proposed a dynamic approach to pseudo-labeling all target data to address incorrect pseudo labels. However, this method does not fully cover all pseudo labels. \cite{jing2024visually} proposed adversarial style matching. In this approach, the goal is to generate source style samples using auxiliary knowledge stored in the pre trained model. However, the lack of anchors, such as prototypes, can degrade the effectiveness of pseudo-labeling in this method.\\
\indent In this study, we explore source-free domain adaptation, where labeled data from the source domain is unavailable. Instead, we rely on the embedded knowledge of a pre-trained model to address the gap between the source and target domains, keeping all the weights in the pre-trained model frozen. However, two main challenges in this context, which have not been adequately addressed in previous works, are: first, the low quality of prototype samples; and second, the incorrect assignment of pseudo-labels. To tackle these challenges, we propose a method consisting of three main phases: 1) prototype generation, 2) pseudo-label assignment, and 3) noisy label filtering. In the first phase, we proposed a Reliable Sample Memory (RSM) that helps improve the quality of prototypes. In the second phase, we use a self-supervised and pseudo-labeling approach that leverages a pre-trained model, reducing uncertainty through a feature selection process. In the final phase, we apply a method to filter out noisy labels, which further improves the quality of the pseudo-labels. Our study makes the following key contributions:
\begin{enumerate}
\item	We introduce a novel prototype generation mechanism driven by self-entropy minimization, which is further enhanced by the proposed Reliable Sample Mining (RSM) strategy. This approach significantly improves the quality of the generated prototypes by emphasizing high-confidence samples, effectively reducing noise in the representation space.
\item We propose a pseudo-label assignment strategy that leverages Multi-View Contrastive Learning, where diverse augmented views of the same data instance are used to enforce representation consistency. This not only mitigates the impact of noisy pseudo-labels but also reinforces semantic alignment across views.
\item Extensive experiments on three widely-used domain adaptation benchmarks demonstrate the effectiveness of our method, consistently outperforming state-of-the-art source-free domain adaptation approaches in terms of accuracy and robustness.
\end{enumerate}
The organization of the paper is as follows: Section 2 provides comprehensive coverage of related works, offering insights into existing SFUDA models. Section 3 outlines the methodology of our proposed model and the corresponding algorithm. In Section 4, we explore the experimental aspects, encompassing evaluation metrics, technical details of the experimental setup, and detailed descriptions of the datasets used. Section 5 concludes with a discussion of the proposed model. Furthermore, in the final section, we discuss the future directions of the SFUDA approach and suggest potential new studies.
\section {Related works} \vspace{0.5cm}
\subsection{traditional domain adaptation} \vspace{0.5cm}
\indent After 2012, the outstanding performance of deep learning in image classification tasks triggered a new era in machine learning. However,  One of the challenging issues that arose is generalization. Unsupervised Domain adaptation(UDA) is a promising approach to tackle this problem. Traditional UDA can be divided into three main approaches. The first, \textbf{domain-invariant features}, uses feature learning methods for the source and target domains to learn shared features. This approach aims to mitigate the domain shift between source and target by extracting common features. Some frameworks benefit from samples and person individual characteristics \cite{li2018domain, lin2020multi}. For instance, progressive disentanglement layers in CNN architectures can separate samples into invariant and specific features \cite{wu2021instance}. Furthermore, studies have utilized person attributes as domain invariant features, showcasing their potential for tasks such as person re-identification in video surveillance \cite{li2020attribute, jia2023domain}.  \\
\indent One common approach in UDA is the \textbf{distance-based method}, which measures the distribution differences between the source and target data using metrics like Maximum Mean Discrepancy (MMD) \cite{gretton2012kernel}. MMD has been a widely used approach \cite{li2020intelligent,rahman2023semi, chen2020homm, zhang2018unsupervised} . Despite its popularity, various modified versions of MMD have been proposed to further mitigate domain shift in specific scenarios. For instance, Joint Maximum Mean Discrepancy (JMMD) \cite{long2017deep} employs a joint distribution that combines source and target data, capturing both domain and class information. However, JMMD might not fully exploit the discriminative information between different classes. Another method, DJP-MMD \cite{zhang2020discriminative}, also adopted a joint distribution but emphasizes inter-class discriminability by placing more weight on pairs of data points from different classes. That said, it might be more sensitive to noise or outliers. The MK-MMD metric, which is employed in the Domain Adversarial Network (DAN), enhances flexibility and adaptability in complex data distributions by leveraging different kernels \cite{long2015learning}.  \textbf{Adversarial domain adaptation}, aims to identify common features by minimizing the importance of discriminative features, employs gradient reversal layer \cite{ganin2016domain} as a key component to enable adversarial training. In DM-ADA method, instead of hard- classifying the data as a real of fake from generated images, it uses soft labeling and can be implemented at both the pixel and feature levels \cite{xu2020adversarial} . \\
\indent Some methods utilize Generative Adversarial Networks (GAN) \cite{tang2020discriminative, xie2020mi, iacono2024structure, fukushi2024few, baek2020weakly}. which can suffer from mode collapse (where the generated data samples lack diversity). Although researches have addressed this issue, a complete solution remains elusive. While traditional UDA has demonstrated promising results, real-world accessibility to training data (source domain) is often limited due to data privacy concerns. To address this challenge, this paper will explore source-free Unsupervised Domain Adaptation (SFUDA) methods that adopt pre-trained models.\\
\indent \textbf{Source-free approach}, One of the promising methods in SFUDA is pseudo-labeling. Different kinds of pseudo-labeling could be utilized in this manner. For instance, nearest neighbors as a clustering method can be used in a geometry-based model, as applied in this study \cite{tang2021nearest}. In this research[15], utilizing a sample-based framework based on a fixed self-entropy threshold to select reliable samples. However, a fixed threshold will give rise to an issue that we will discuss in the next section. selecting the reliable samples in these methodologies are bottleneck. One of the trending approaches in SFUDA involves the use of self-supervised techniques to pseudo-label pretrained models\cite{song2022ss8}. Leveraging the class prototype concept involves assigning weights to each sample to establish a class prototype. Despite this method employing a novel strategy to tackle the issue of sample selection, it also depends on pseudo-labeling and self-training techniques \cite{zhou2024source}. Vision Transformers have been utilized in numerous research studies to adopt the self-attention mechanism, which is applicable for selecting more reliable features \cite{rizzoli2024source, yang2021transformer, xu2021cdtrans, zhang2023universal}. Based on the gaps in these researches, we propose a three-phase concept of source-free domain adaptation, as shown in Figure 1. In the first phase, prototype generation is based on RSM and self-entropy. In the next phase, we introduce a multi-contrastive learning approach to optimize pseudo-labeling, and in the final phase, we apply noisy label filtering.\\
\begin{figure}
  \includegraphics[width=\linewidth]{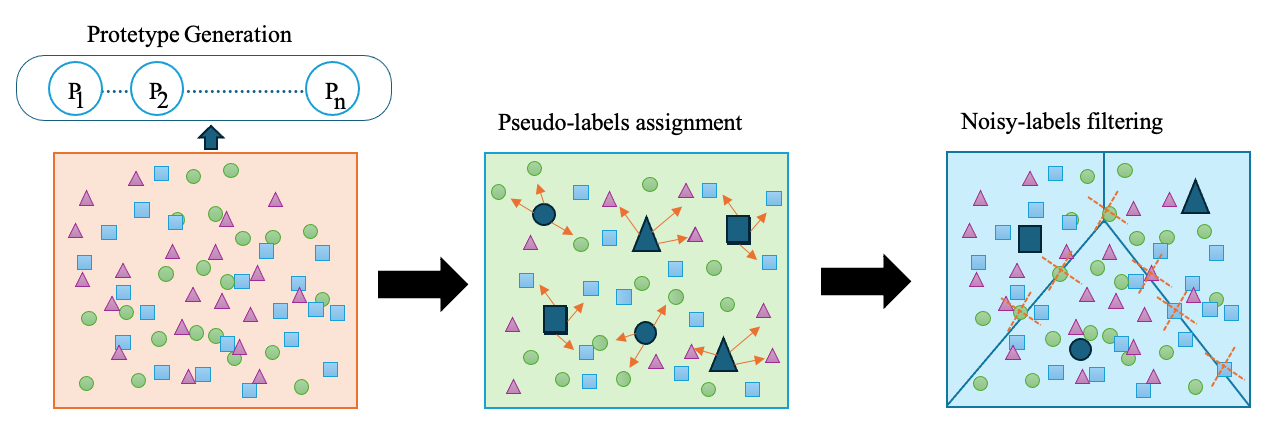}
  \caption{Schematic Representation of the Three Phases of Our Approach}
  \label{fig:fig1}
\end{figure}
\section{Method} \vspace{0.5cm}
\subsection{problem statement} \vspace{0.5cm}
In the context of UDA, $x_s$ and $y_s$ represent the datapoints and labels of the source domain, respectively. In the target domain, we merely have datapoints $x_t$ and the labels are not available. Accordingly, in mathematical form we have
\[
D_S = \{x_s^i, y_s^i\}_{i=1}^{n_s},
\]
for the source domain, and
\[
D_T = \{x_t^j\}_{j=1}^{n_t}
\]
for the target domain. The main concept of the UDA method is that the probability distributions are related but different.

However, in SFUDA, access to source samples is restricted due to data privacy. We have just pre-trained parameters $\theta_s$ and target samples $\{x_j^t\}_{j=1}^{n_t}$.
  \begin{figure}
  \includegraphics[width=\linewidth]{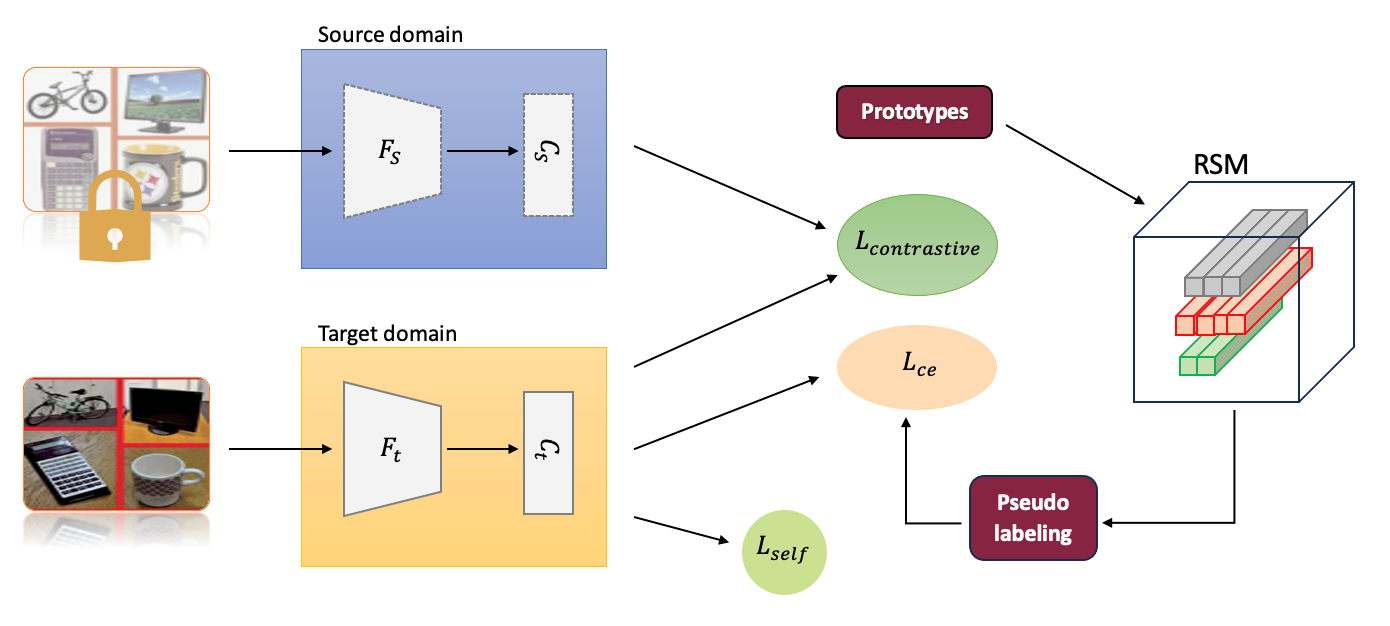}
  \caption{An Overviewe of the proposed method}
  \label{fig:fig2}
\end{figure}
\subsection{Framework}

As shown in Figure~(2), in our proposed framework, two key components are embedded: the first is a pre-trained source model, and the second is the target domain. Within the pre-trained source model, the term $\theta_s$ denotes the fixed parameters of this domain. Additionally, we use $C_s$ to refer to the source classifier and $F_s$ as the feature extractor. As mentioned earlier, the source domain training set is unavailable due to privacy concerns. Therefore, our approach focuses on leveraging knowledge from the pre-trained source model and datapoints from the target domain.

In the pre-trained source model, the parameters of the source classifier are denoted as $\theta_{C_s}$ and $\theta_{F_s}$ for the parameters of the pre-trained source model's feature extractor. Furthermore, all parameters are fixed after pre-training. 

In the target domain, $C_t$ and $F_t$ represent the target classifier and feature extractor, respectively. In our proposed framework, we employed three loss functions: $L_{\text{contrastive}}$, $L_{\text{ce}}$, and $L_{\text{clustering}}$. The RSM plays a crucial role in prototype generation.
\subsection{proptotype generation } \vspace{0.5cm}
In this module, the model identifies reliable samples to represent each class, referred to as anchors. Each class has an anchor, determined by the minimum self-entropy within that class. A straightforward approach is to select a fixed number of samples with low self-entropy for each class. However, variations in entropy distribution across classes may result in some classes having a lower average self-entropy than others. The primary disadvantage of these methods is the potential loss of crucial information. To address this issue, we employ a method that uses a flexible threshold, which can adapt dynamically in each iteration. In the proposed method, we leverage RSM to select prototypes in the prototype generation module. This approach, in each iteration, selects the best-fit prototype to determine the appropriate threshold and chooses the best samples as the output of this module.
\subsubsection{Reliable Sample Memory(RSM)} \vspace{0.5cm}

Reliable samples play a pivotal role in our method. One of the challenges in previous works is selecting reliable samples and leveraging them effectively. In Reliable Sample Memory (RSM), we first calculate the self-entropy of each target sample $\{x_j^t\}_{j=1}^{n_t}$. According to Equation~(\ref{eq:minmax}), we apply min-max normalization to the self-entropy using Equation~(\ref{eq:entropy_normalization}). Next, we sort all values in the matrix $E$ according to Algorithm~(1). 

To begin, we calculate the self-entropy based on Equation~(\ref{eq:entropy_normalization}) for each class, and the minimum entropy of each class is used as its representative value. The $E$ matrix is $M \times N$, where $M$ represents the number of iterations and $N$ is the number of classes. According to Equation~(\ref{eq:minmax}), we maximize the minimum entropy values for each class over $M$ iterations and use this as the threshold:
\begin{equation}
    \eta = \max \left\{\min \left(E_{i,j} \right) \right\}_{i=1}^{{M, j=1}^{N}}.
    \label{eq:minmax}
\end{equation}

Then, all values are sorted in ascending order from row 1 to $n$. One advantage of representing data in matrix form is that matrices provide a compact and structured way to represent data. Additionally, the use of matrix forms allows for more convenient batch processing of data and helps to minimize the computational cost in our model as well.

The self-entropy is defined as:
\begin{equation}
    H(p) = -\sum_i p_i \log p_i,
    \label{eq:entropy}
\end{equation}

and after normalization:
\begin{equation}
    H(x') = -\sum_i \frac{x_i - \min(x)}{\max(x) - \min(x)} \log \left(\frac{x_i - \min(x)}{\max(x) - \min(x)}\right).
    \label{eq:entropy_normalization}
\end{equation}

Based on Algorithm~(2), we addressed the selection of reliable  samples, for each class. A straightforward method might involve choosing a fixed number of samples per class based on the lowest self entropy. However, such an approach fails to account for the varying entropy distributions across different classes, which can lead to biases in prototype selection due to lower average self entropies in some classes. After calculating $c_k$, which represents the prototypes of the distribution of each category in the target domain, we assign pseudo labels, which are discussed further in the next section. 
  \begin{figure}
  \includegraphics[width=\linewidth]{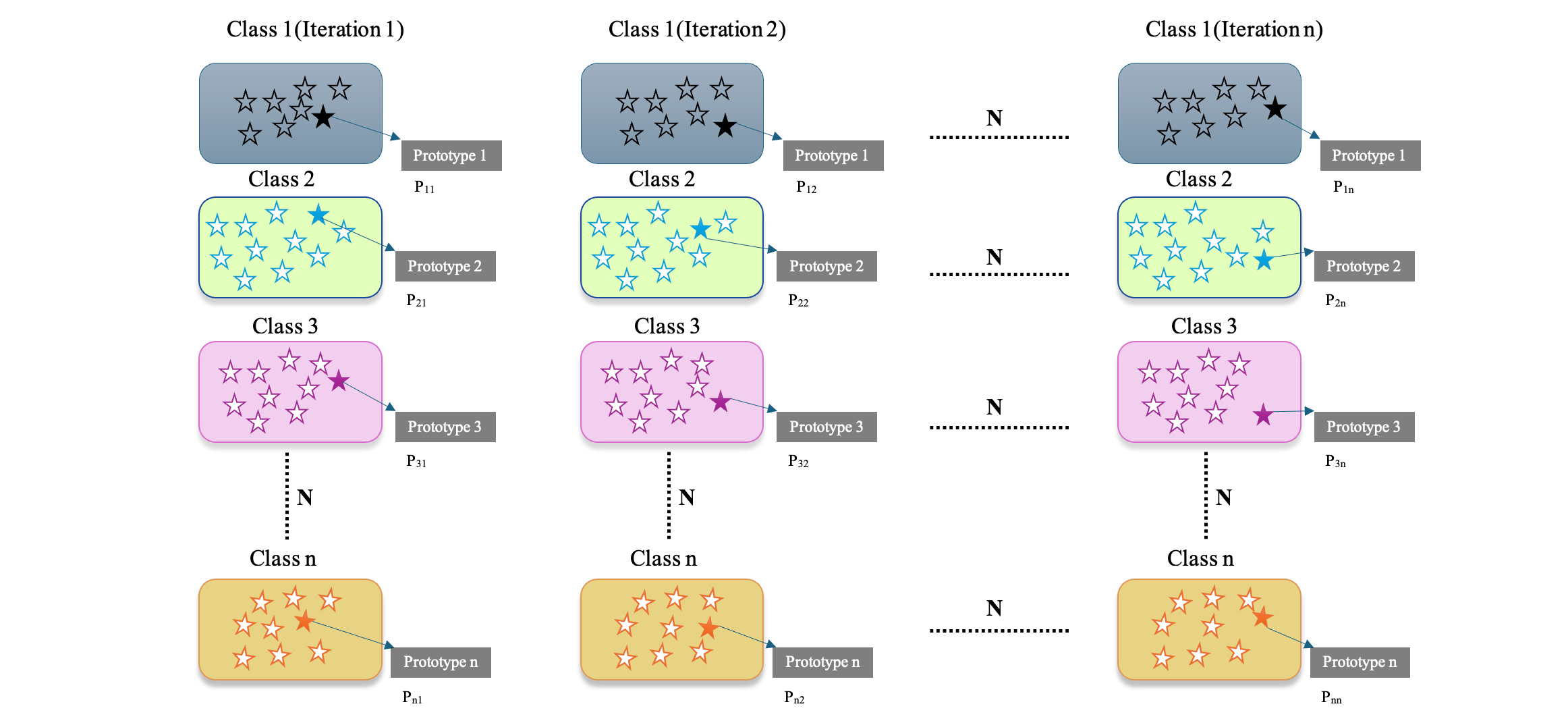}
  \caption{The prototype generation phase based on RMS and self entropy }
  \label{fig:fig3}
\end{figure}
Figure 3 provides a clear visualization of how the Reliable Sample Memory (RSM) method works in practice. At first, each class includes samples with different entropy levels, shown here by stars darker stars mean more reliable samples due to their lower entropy. As the method proceeds through multiple iterations, RSM continually refines the choice of class prototypes by selecting the most reliable samples based on minimum entropy thresholds. This iterative process helps address biases that could occur if some classes naturally had more confident predictions than others. Ultimately, this makes the chosen prototypes more accurate and trustworthy, which directly improves the quality of pseudo-label assignments for further analysis.

\begin{algorithm}
\caption{Reliable Sample Memory (RSM)}
 \textbf{Input:} Target samples $D_t = \{x_i^j\}_{j=1}^{N_t}$, number of iterations $M$, number of classes $N$\\
 \textbf{Output:} Threshold $\eta$\\
\textbf{begin} \\
For {$c \leftarrow 1$ to $N$}\\
      $  H_c \leftarrow []$ \texttt{//}
Initialize class-wise entropy set\\
End For\\
For {$t \leftarrow 1 $  to $N_t$}\\
     $  (\hat{y}_t, H) \leftarrow F_t(C_t, x_t)$ \texttt{//}
 Compute predicted label and entropy \\
     $  H_{\hat{y}_t} \text{.append}(H)$ \texttt{//}
 Append entropy to the corresponding class\\
End For\\
For {$c \leftarrow 1 $ to $N$}\\
     $  H_c \leftarrow \text{MinMaxNormalize}(H_c)$ \texttt{//}
 Normalize entropy values for class  $c$\\
End For \\
 $E \leftarrow [H_1, H_2, \ldots, H_N]$ \texttt{//}
 Combine normalized entropies into matrix $E$\\
For {$i \leftarrow 1$ to $M$} \\
     $  \text{minEntropy}[i] \leftarrow \min\{E[i, j]\}$ for $j \leftarrow 1$ to $N$ \\
End For \\
 $\eta \leftarrow \max\{\text{minEntropy}[i]\}$ for $i \leftarrow 1$ to $M$ \\
 \textbf{return} $\eta$ \\
 \textbf{end} \\
\end{algorithm}
Algorithm 2 outlines the initialization and updating process of the Reliable Sample Memory (RSM) module. The procedure begins by computing the self-entropy values for each sample using Equation (2), followed by sorting these values in ascending order, such that the most reliable (i.e., lowest entropy) samples appear first. These sorted entropy values are then organized into a matrix structure, enabling balanced representation across all classes. By partitioning the data into segments of equal size, the algorithm facilitates efficient management of reliable samples throughout the iterative refinement process. This structured selection mechanism enhances the robustness of prototype construction and reduces potential bias during pseudo-label assignment.

\begin{algorithm}
    \caption{Initialize/Update}
    \label{alg:initialize_update}
         \textbf{Input:} $n^2$ input from eq(2) \\
         \textbf{Output:} Reliable Sample Memory $R$ (RSM) \\
         begin \\
         Initialize entropy\_values[$N$] \\
        For {$i = 0$ \textbf{to} $N-1$} \\
             entropy\_values[$i$] = $-\sum_{j=0}^{C-1} (probabilities[i][j] \cdot \log(probabilities[i][j]))$ \\
        End For \\
         Sort entropy\_values in ascending order \\
         Define row\_size = $N / n$ \\
         Initialize matrix $M$ with dimensions $[n][row\_size]$ \\
        For {$k = 0$ \textbf{to} $n-1$} \\
             Start\_index = $k \cdot row\_size$ \\
             End\_index = Start\_index + row\_size \\
            For {$m = Start\_index$ \textbf{to} End\_index-1}  \\
                 $M[k][m - Start\_index] = entropy\_values[m]$ \\
            End For \\
        End For \\
         return $M$ \\
         end \\
\end{algorithm}
\subsection{ Pseudo label assignment based (Multi-View contrastive learning)} \vspace{0.5cm} 
Pseudo-labeling generates supervisory signals for the target domain using the pre-trained source model, which serve as a proxy for ground truth, facilitating supervised learning in the target domain. These pseudo-labels are refined iteratively to align more closely with the data distribution.

In this case, the pseudo-label is directly assigned to the category for which the model outputs the highest confidence. 

In the proposed method, Multi-view contrastive learning (MVCL) and pseudo labeling are utilized. In this approach, we use data augmentation to generate samples, which are then used to set different features $\{x^{(v)}\}_{v=1}^V$.

In the proposed method, we address this issue through contrastive learning, as explained in the following sections. Initially, data augmentation techniques, including color jittering and rotation, are employed to enhance the dataset. Let $D = \{x_i, y_i\}_{i=1}^N$ be a dataset, where $x_i$ represents the input data and $y_i$ represents the corresponding label. The dataset $D$ and $N$ represent the number of samples. The transformation function $T: x \longrightarrow x'$ applied to the input data where $x$ is the input space. The augmented dataset is $D' = \{T(x_i), y_i\}_{i=1}^N$ where $T(x_i) = \{T_j(x_i)\}_{j=1}^M$.

In this equation, $T_j$ represents different augmentation operations, and $M$ is the number of different transformations applied to each sample. Feature extraction involves applying a feature extraction function $\phi: X \rightarrow F$ to the augmented data, where $F$ is the feature space. Here, $f_{i,j} \in F$ represents the feature vector extracted from the $j$-th augmented version of the $i$-th original sample.

\begin{equation}
f_{i,j} = \phi(T_j(x_i)) \quad \text{for} \quad i = 1, \ldots, N \quad \text{and} \quad j = 1, \ldots, M.
\label{eq:4}
\end{equation}

After augmentation and feature extraction, feature concatenation is employed. Due to the different views, and in order to align the features, we applied weighted feature concatenation based on a data-driven method. More precisely, weights are assigned to each view to reflect their relative importance. In our method, we use the variance of the feature set to measure the discrimination levels. Crucially, datasets with higher variance exhibit more dispersion, indicating the need for more diverse and relevant patterns for clustering.

Let $n$ views and $F_i$ be the feature matrix corresponding to the $i$-th view of $m$ samples and $d_i$ features, thus the dimension is $m \times d_i$. The variance $\sigma_i^2$ for the $i$-th view is calculated from equation \eqref{eq:5} as the average of variance of each feature:

\begin{equation}
\sigma_i^2 = \frac{1}{d_i} \sum_{j=1}^{d_i} \text{Var}(F_{i,j})
\label{eq:5}
\end{equation}

where $F_{i,j}$ represents features in the $i$-th view and $j$-th feature, and $\text{Var}$ is the variance of features throughout $m$ samples. To compute $w_i$ for the $i$-th view, the variance is normalized throughout all views based on equation \eqref{eq:6}.

\begin{equation}
w_i = \frac{\sigma_i^2}{\sum_{k=1}^{n} \sigma_k^2}
\label{eq:6}
\end{equation}

This normalization can improve the interpretability of the weights and facilitate the process. It also regularizes the contribution of each view and prevents one view from dominating the others. After concatenating features, the multi-view clustering problem essentially transforms into a single-view clustering task. This transformation simplifies the overall approach, enabling us to apply standard clustering algorithms effectively. In our study, we adopt K-means for clustering due to its simplicity and well-established performance. The K-means algorithm iteratively partitions the data into $k$-clusters by minimizing the within-cluster variance, which is particularly suitable for handling the combined feature. The feature vectors from all views for a given sample are concatenated into one composite representation, with each view’s feature scaled by a weight reflecting its importance as shown in equation \eqref{eq:7}.

\begin{equation}
z_i = \begin{bmatrix} 
w_1 h_i^{(1)} \\ 
w_2 h_i^{(2)} \\ 
\vdots \\ 
w_v h_i^{(V)} 
\end{bmatrix}
\label{eq:7}
\end{equation}

where $z_i$ denotes concatenation and $w_i$ is the weight for view $v$, and $h_i$ denotes a vector of dimension of features.

MVCL aims to partition data across multiple views. Using MVCL, the model leverages complementary information, offering an advantage over traditional clustering methods. In the proposed method, an attention-based mechanism is employed to fuse the features effectively. A major flaw in previous methods involving pseudo-labeling is their inability to properly balance categories during the pseudo-labeling phase, which is a crucial aspect of this approach. To address this issue, we proposed an attention-based mechanism. In equation \eqref{eq:8}, $w_{i,j}$ represents the attention weights between samples $i$ and $j$. In equation \ref{eq:8}, $z_i$ and $z_j$ represent a positive pair, and $z_i$ and $z_k$ represent a negative pair. ‘$\text{Sim}$’ represents the similarity function in the mentioned equation. $\tau$ is a temperature parameter that scales the similarity. $1[k \neq i]$ means that if $k \neq i$, this coefficient equals one and cancels out.

\begin{align}
l_{\text{contrastive}} &= \sum_i \left[ w_{i,j} - \log \frac{\exp(\text{Sim}(z_i, z_j) / \tau)}{\sum_{k=1}^{2N} 1[k \neq i] \exp(\text{Sim}(z_i, z_k) / \tau)} \right] 
\label{eq:8}
\end{align}

Equation \eqref{eq:9} represents the cluster loss function, where $x_i$ refers to the data points, $\mu_j$ is the centroid of the cluster, and $n$ is the number of points in the cluster. This formula computes the squared differences between the data points and their corresponding centroids, which is minimized to achieve better clustering.

\begin{equation}
l_{\text{cluster}} = \sum_{i=1}^{n} \min \|x_i - \mu_j\|^2
\label{eq:9}
\end{equation}

In equation \eqref{eq:11}, calculate cluster loss where $\mu_j$ is the cluster centroid and $x_i$ is the data point. In equation \ref{eq:10}, $l_{\text{contrastive}}$ is the contrastive loss, and $l_{\text{cluster}}$ is the clustering loss. In this equation, $L_{\text{ce}}$ represents the Cross-entropy loss for pseudo-labeled target samples; and Contrastive loss ensures consistent feature representations. In the equation, $\lambda_1$, $\lambda_2$, and $\lambda_3$ are weighting factors. Furthermore, $\lambda_1$, $\lambda_2$, and $\lambda_3$ are the trade-off parameters which control the relative contribution of contrastive and clustering losses.

\begin{align}
l_{\text{total}} &= \lambda_1 l_{\text{contrastive}} + \lambda_2 L_{\text{ce}} + \lambda_3 l_{\text{clustering}}  \nonumber \\
\text{s.t.} \quad \sum_{i=1}^{3} \lambda_i &= 1
\label{eq:10}
\end{align}

 \begin{figure}
 \centering
  \includegraphics[width=0.5 \linewidth]{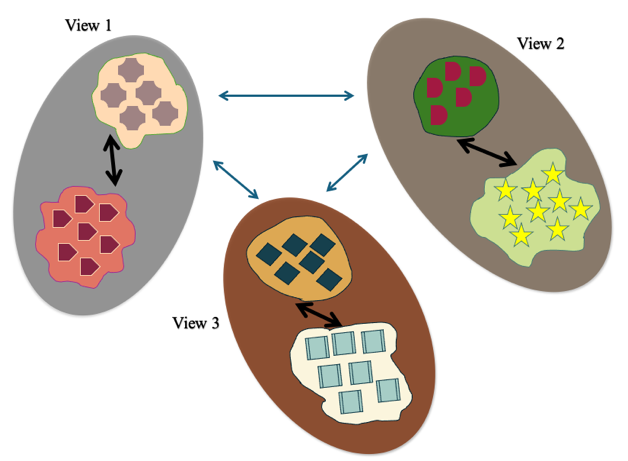}
  \caption{schematic of multi-view  approach. }
  \label{fig:fig4}
\end{figure}

\subsection{Pseudo-labels filtering}  \vspace{0.5cm}

Due to domain discrepancies, noisy and unreliable labels often appear in the model's output, reducing accuracy and presenting a common challenge in pseudo-labeling. To mitigate this, we employ an adaptive threshold instead of a fixed threshold throughout the training process. This adaptive threshold allows the model to include more data when uncertainty is high and to be more selective as confidence increases. In our approach, we first compute the self-entropy for each sample $x_t$ and collect these values into a set for each class. Instead of using a fixed threshold, we define, as in equation \eqref{eq:11}, an adaptive threshold $\theta$ as the maximum of the minimum entropies across all classes:

\begin{equation}
\theta = \max \{ \min H_c \mid c \in C \}
\label{eq:11}
\end{equation}

where $H_c$ represents the entropy set for class $c$ and $C$ denotes the set of all classes. As training progresses, this threshold $\theta$ dynamically decreases, allowing the model to focus on more confident and reliable samples as it becomes more accurate. In each iteration, we compute $\theta_1, \theta_2, \theta_3, \ldots, \theta_n$, where $n$ is the total number of iterations, by determining the maximum of the minimum entropy values for each class. We then use an attention mechanism over $[\theta_1, \theta_2, \theta_3, \ldots, \theta_n]$ and assign corresponding attention weights $[\propto_1, \propto_2, \propto_3, \ldots, \propto_n]$. The attention scores $S_i$, as in equation \eqref{eq:12}, are computed using the standard dot product attention mechanism:

\begin{equation}
S_i = \frac{\theta^T e_i}{\sqrt{d}}
\label{eq:12}
\end{equation}

where $\theta^T$ is the transposed threshold matrix, $e_i$ is the entry, and $d$ represents the dimensionality of embedding vectors or the scaling factor. Finally, we compute the weighted average threshold across $\rho$ iterations using, as in equation \eqref{eq:13}. This formulation ensures that the pseudo-label selection process dynamically adapts to the confidence levels of the model, improving the reliability of the selected samples.

\begin{equation}
\frac{1}{\rho} \sum_{i=1}^{\rho} \propto_i \cdot \theta_i
\label{eq:13}
\end{equation}

\section{Experiments}

\subsection{Datasets and Experimental Setup} \vspace{0.5cm}

In this section, we describe the three widely adopted benchmark datasets commonly used in the SFUDA research: VisDA-2017 \cite{xie2020mi}, Office-31 \cite{fukushi2024few}, and Office-Home \cite{iacono2024structure}. We also detail the experimental configuration and evaluate our approach against current state-of-the-art SFUDA methods. Additionally, we conduct an ablation study to assess the impact of each individual component in our proposed framework.

\subsection{Benchmark Datasets} \vspace{0.5cm}

\textbf{VisDA-2017}: This benchmark addresses a challenging scenario of adapting from synthetic to real images across 12 categories. It comprises approximately 152,000 synthetic images for the source domain and 55,000 real-world images in the target domain, providing a rigorous evaluation setup for domain adaptation.

\textbf{Office-31}: A standard benchmark in domain adaptation, Office-31 includes 31 categories of office-related items collected from three distinct domains: Amazon (A), DSLR (D), and Webcam (W). The domains contain 2,817, 498, and 795 images respectively, enabling a diverse assessment across different environments.

\textbf{Office-Home}: This dataset spans 65 object categories across four domains: Art (A), Clipart    (C), Product (P), and Real-World (RW). With its increased variability and complexity, it poses a more difficult challenge for domain adaptation techniques.

\subsection{Evaluation Metrics} \vspace{0.5cm}

To evaluate model performance, we adopt the standard metric commonly used in source-free domain adaptation: classification accuracy on the target domain. As shown in Equation (14), this is calculated by checking how often the predicted labels (\( \hat{y}_t \)) match the ground-truth labels (\( y_t \)) for each sample (\( x \)) in the target set (\( X_t \)). The accuracy is computed using the following formula:

\begin{equation}
\mathrm{Accuracy} = \frac{\left|\{x \mid x \in X_t \land \hat{y}_t = y_t\}\right|}{\left|\{x \mid x \in X_t\}\right|}
\label{eq:14}
\end{equation}

Here, \( x \in X_t \) refers to a target domain sample, \( y_t \) is its ground-truth label, and \( \hat{y}_t \) is the predicted label. This metric provides a simple yet effective way to compare the performance of SFDA methods.

\subsection{ Implementation details} \vspace{0.5cm}

The experimental framework is developed using the PyTorch deep learning library. Prior to training, all input images are subjected to a preprocessing pipeline involving random resizing and cropping to a resolution of $256 \times 256$ pixels. Subsequently, two augmentation strategies are applied: (1) a mild augmentation through random horizontal flipping, and (2) a stronger augmentation that incorporates random rotations, brightness variations, and scaling transformations, in line with established practices such as those in \cite{zhang2018unsupervised}. For the backbone architecture, we adopt the ResNet-101 model pre-trained on the source domain, consistent with previous works \cite{liu2025adaptive}. To enhance adaptability to the target domain, the original fully connected (FC) layer is substituted with a task-specific FC layer followed by a batch normalization (BN) layer. The domain discriminator employed in our SFDA framework comprises a three-layer fully connected network, with hidden units set to 512, 1024, and 2048, respectively. Model optimization is carried out using the Adam optimizer, with learning rates configured at 0.001 and 0.0005. The training process is conducted over 100 epochs for VisDA-2017 and extended to 150 epochs for the Office-31 and Office-Home benchmarks. A batch size of 32 is utilized across all tasks to balance computational efficiency and memory usage.

We use a batch size of 32 for all tasks, which ensures efficient training without overloading memory. To enable end-to-end optimization of $F_s$, $C_s$, and $D$, we incorporate the gradient reversal layer (GRL) \cite{ganin2015unsupervised} during training. Regarding the trade-off of hyperparameters in equation \ref{eq:10},  

\begin{figure}
  \includegraphics[width=\linewidth]{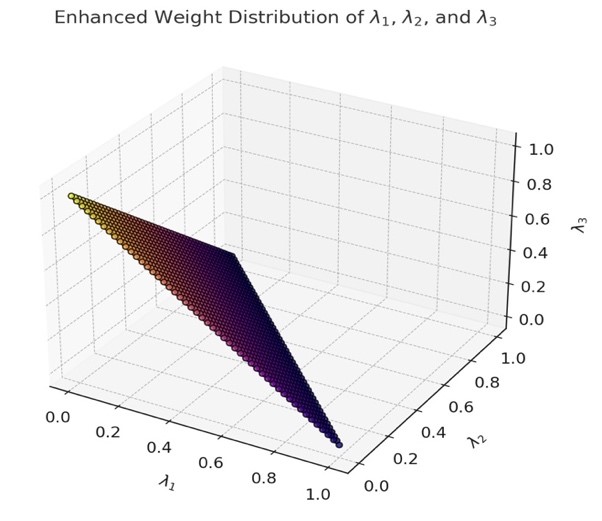}
  \caption{Change of Hyperparameters ($\lambda_1, \lambda_2$ and $\lambda_3$).
 }
  \label{fig:fig5}
\end{figure}
\begin{figure}
  \includegraphics[width=\linewidth]{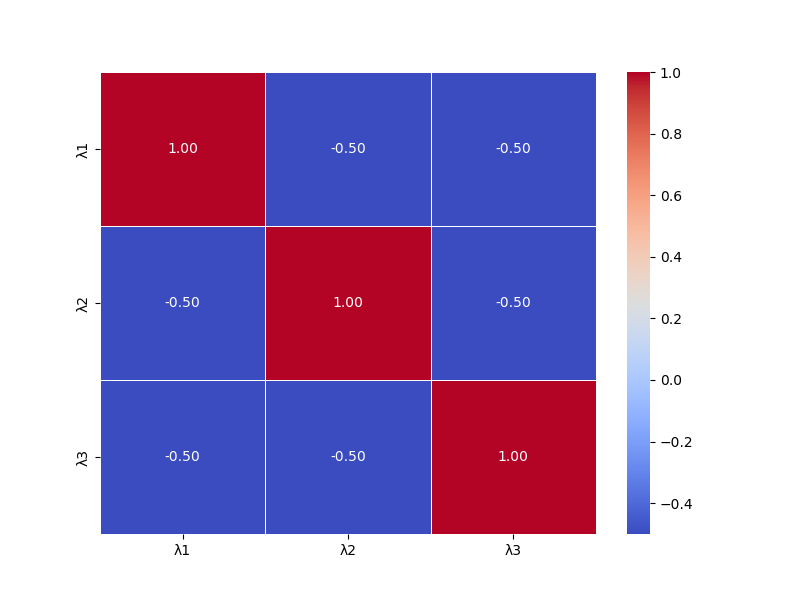}
  \caption{ Correlation Matrix of Hyperparameters ($\lambda_1$, $\lambda_2$, $\lambda_3$) }
  \label{fig:fig6}
\end{figure}
We acknowledge that at the start of training, the model is more inclined to learn simple patterns. As a result, it benefits from stronger supervision provided by pseudo-labels in the early stages. However, as training advances, the potential for incorrect pseudo-labels increases, which can negatively affect the model’s learning process. To address this, we gradually reduce the model’s dependence on these pseudo-labels and place more emphasis on the distribution consistency constraint, ensuring that the network is less misled by erroneous labels as it learns.

\subsection{Sensitivity Analysis of Parameters}  \vspace{0.5cm}

The 3D scatter plot provides a clear visualization of how the weighting factors $\lambda_1$, $\lambda_2$, and $\lambda_3$ are distributed while ensuring they sum to one. The color gradient helps illustrate how $\lambda_3$ changes in response to variations in $\lambda_1$ and $\lambda_2$. As expected, when $\lambda_1$ and $\lambda_2$ increase, $\lambda_3$ decreases, highlighting the trade-offs in weight allocation. This emphasizes the importance of maintaining a well-balanced distribution; if one weight becomes too dominant, the others lose their effectiveness.

This balanced weight distribution is crucial in loss function optimization, as it determines the relative contributions of contrastive loss, cross-entropy loss, and clustering loss in the objective function. The smooth transition in $\lambda_3$ values across the plot indicates a continuous and systematic trade-off between these terms, which can be optimized based on empirical performance. The correlation matrix gives us a clearer picture of how the weights $\lambda_1$, $\lambda_2$, and $\lambda_3$ interact. As expected, each weight is perfectly correlated with itself (1.0, 1.0, 1.0), but what's more interesting is how they relate to each other. $\lambda_1$, $\lambda_2$, and $\lambda_3$ have a correlation of $-0.5$ to $-0.5$, meaning that increasing one tends to reduce the other, but not in a strictly opposite way. The same pattern appears between $\lambda_1$ and $\lambda_2$, as well as $\lambda_2$ and $\lambda_3$, showing that the three weights are closely tied together due to the constraint that their sum must always be 1.

\begin{figure}
  \includegraphics[width=\linewidth]{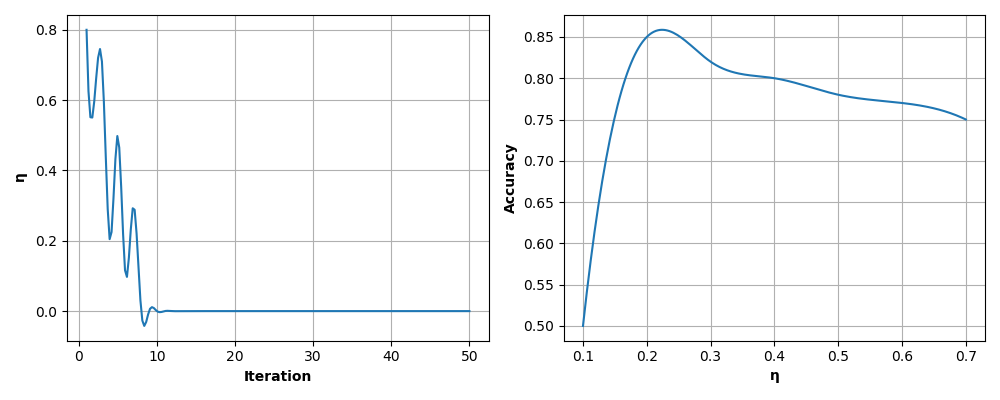}
  \caption{Change of $\eta$ according to the number of training iterations (left). Performance change with regard to fix entropy threshold $\eta$ (right)  }
  \label{fig:fig7}
\end{figure}

In our research, the parameter $\eta$ plays a critical role in identifying reliable samples within the Reliable Sample Memory (RSM) framework. As illustrated in the left chart of Figure \ref{fig:fig7}, $\eta$ exhibits substantial fluctuations during the initial iterations, ranging from approximately 0.1 to 0.8. This instability suggests significant variation in entropy values among different classes, making it challenging to identify truly reliable samples. However, after approximately 20 iterations, $\eta$ stabilizes near zero, indicating that the model has effectively filtered out uncertain samples and converged on a more consistent set of reliable ones. The rapid stabilization of $\eta$ underscores the efficiency of the entropy-based selection process in refining the sample pool during the early stages of the learning phase.

The right chart of Figure \ref{fig:fig7} demonstrates the influence of $\eta$ on model accuracy. As $\eta$ increases from 0.1 to approximately 0.3, accuracy progressively improves, reaching a peak of approximately 85\% when $\eta$ falls within the 0.3–0.35 range. However, as $\eta$ continues to increase toward 0.7, accuracy declines. This pattern aligns with expectations—assigning a low value to $\eta$ may exclude valuable samples, whereas a higher value introduces unreliable, high-entropy samples that adversely affect model performance. The observed trend suggests that $\eta$ must be carefully calibrated to balance the exclusion of noise while preserving informative data.

The considerable variability in $\eta$ during the early iterations indicates significant divergence in entropy calculations across different classes. The subsequent stabilization suggests that the model systematically refines its selection criteria, effectively eliminating unreliable samples. This progression highlights the robustness of the entropy-based filtering mechanism, ensuring that the model remains resilient against high-entropy noise. Furthermore, the near-parabolic shape of the accuracy curve reaffirms the necessity of optimizing $\eta$; excessively low values restrict learning, whereas excessively high values introduce substantial uncertainty, both of which lead to suboptimal performance.

\subsection{Visual Analysis} \vspace{0.5cm}

\begin{figure}
  \includegraphics[width=\linewidth]{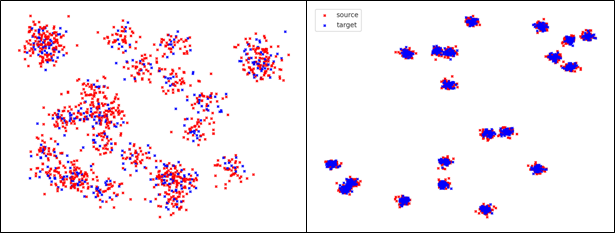}
  \caption{ feature space visualization (left) and the proposed method (right)  }
  \label{fig:fig8}
\end{figure}

Figure \ref{fig:fig8} illustrates the distribution of features from the source (red) and target (blue) domains before applying our unsupervised domain adaptation (UDA) method. The two domains appear clearly misaligned. A model trained on the source domain would likely struggle to generalize well.

To the target domain due to these differences in feature space. In contrast, Figure \ref{fig:fig8}-right shows the feature distributions after applying our method. The source and target data points now form more cohesive and dense clusters, indicating that the domain shift has been significantly reduced. This alignment suggests that our approach effectively learns a domain-invariant feature representation, allowing for better generalization across domains.

To get a clearer picture of how these methods compare, we looked at three key metrics: labeling rate (Ratio 1), model accuracy (Ratio 2), and labeling accuracy (Ratio 3). SHOT++ performs reasonably well, keeping model accuracy around 82.9\% and labeling accuracy at 78.9\%, though 

\begin{figure}
  \includegraphics[width=\linewidth]{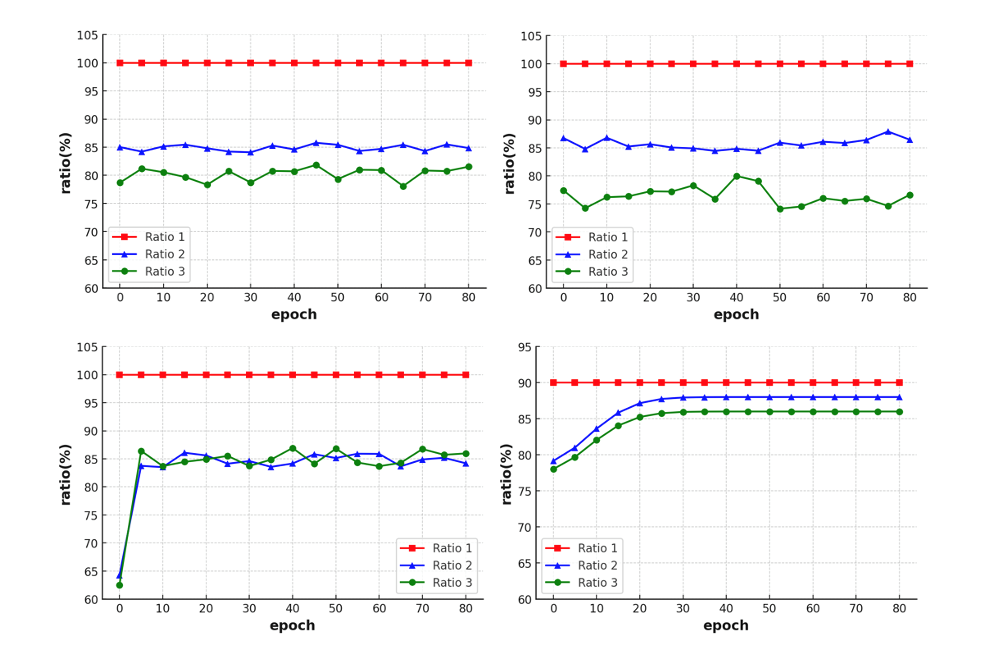}
  \caption{Progression of pseudo labeling during adaptation. The red curve indicates the labeling rate, the green curve shows pseudo label accuracy, and the blue curve represents overall model accuracy.
  }
  \label{fig:fig9}
\end{figure}
There are slight fluctuations (standard deviation $\approx$ 1.1\%). SHOT, on the other hand, performs slightly worse, with model accuracy dropping to 81.9\% and labeling accuracy to 74.6\%, showing even more instability (std. dev $\approx$ 1.4\% for labeling accuracy). SFDA exhibits a rapid initial improvement in accuracy; however, it suffers from significant instability, as indicated by a high standard deviation (5.3\% for model accuracy). While it does reach 89\% model accuracy and 87\% labeling accuracy in the end, its unpredictable jumps make it less reliable in real-world applications. By contrast, the proposed method consistently outperforms all compared approaches. Both model and labeling accuracies show a stable and continuous improvement, ultimately converging at 90\% and 88\%, respectively, with minimal variance. This smooth and reliable optimization behavior, coupled with the highest final accuracy, underscores the effectiveness and robustness of our approach in domain adaptation tasks.

\subsection{Ablation Study} \vspace{0.5cm}

In this ablation study, we analyze three key components of the framework: Prototype Generation (PG), Pseudo-label Assignment (PA), and Noisy-label Filtering (NF), as shown in Table~\ref{tab:ablation}. The study evaluates the model's performance across four domain adaptation tasks (A$\rightarrow$D, A$\rightarrow$W, D$\rightarrow$A, and W$\rightarrow$A) using three datasets: VisDA-2017, Office-31, and Office-Home. 

The base model (direct transfer) performs with accuracy ranging from 61.8\% to 69.4\%, showing the limitations of a simple transfer learning approach. Incorporating PA improves performance significantly, with accuracy increasing by 23.7\% to 27.6\%, especially in the A$\rightarrow$D task, where accuracy reaches 92.1\%. When combined with PLA, the model's accuracy rises further by 3.7\% to 4.8\%, particularly excelling in tasks like A$\rightarrow$D and A$\rightarrow$W.

The most significant performance boost is observed when NF is integrated alongside PA and PLA. Accuracy increases by an additional 3.4\% to 4.0\%, culminating in an impressive 98.5\% on the A$\rightarrow$D task. These results clearly demonstrate that the joint use of PA, PLA, and NF is crucial for achieving optimal performance. Specifically, NF plays an essential role in mitigating the impact of noisy labels, thereby enhancing the model’s generalization capabilities across diverse domain adaptation scenarios.
\begin{table}[h!]
    \centering
    \caption{Performance comparison across ablation study components}
    \label{tab:ablation}
    \begin{tabular}{@{}lcccc@{}}
        \toprule
        Task & Base Model & +PA & +PA+PLA & +PA+PLA+NF \\ 
        \midrule
        A$\rightarrow$D & 69.4 & 92.1 & 96.8 & \textbf{98.5} \\ 
        A$\rightarrow$W & 61.8 & 85.5 & 90.3 & \textbf{94.3} \\ 
        D$\rightarrow$A & 65.2 & 88.9 & 93.7 & \textbf{97.7} \\ 
        W$\rightarrow$A & 68.0 & 91.6 & 95.2 & \textbf{98.0} \\ 
        \bottomrule
    \end{tabular}
\end{table}
\begin{figure}
  \includegraphics[width=\linewidth]{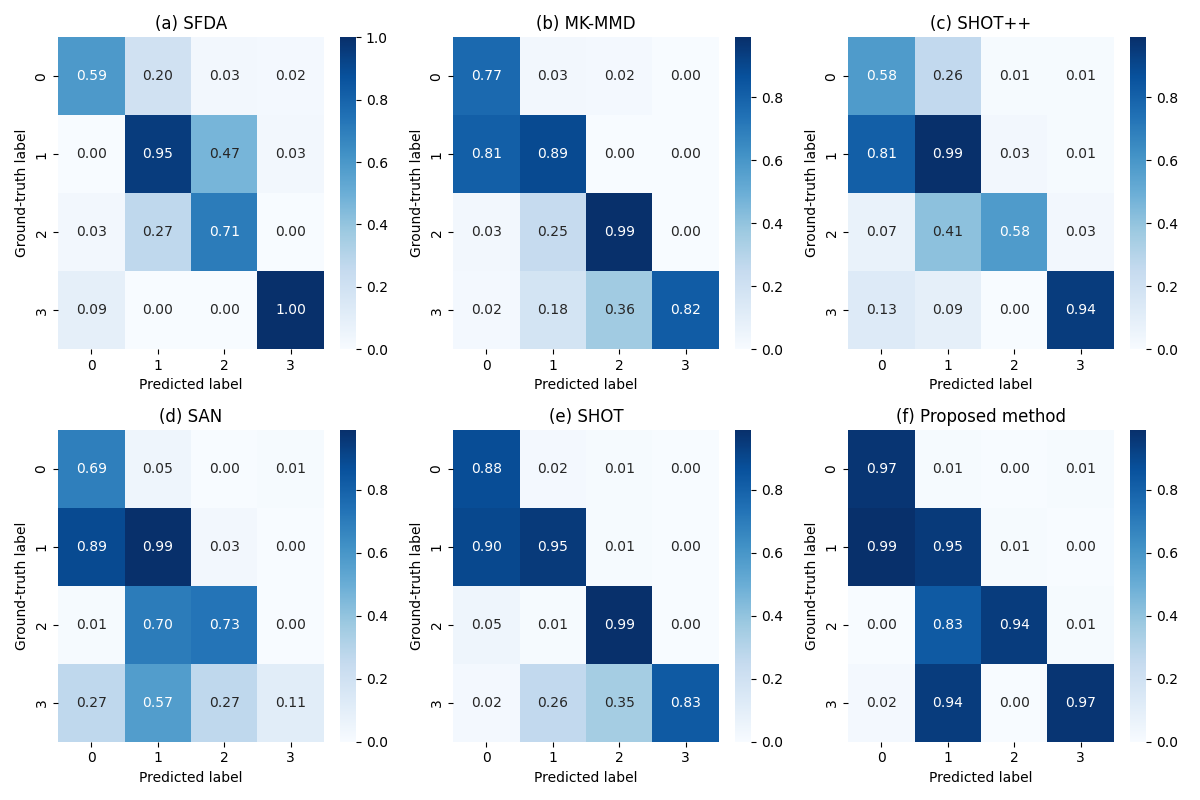}
  \caption{Confusion matrix of five state-of-the-art methods compared with the proposed method }
  \label{fig:fig10}
\end{figure}
In this study, as shown in Figure~(10), we compared the performance of six different models across four domain adaptation tasks: SFDA, MK-MMD \cite{su2024two}, SHOT++ \cite{liang2021source}, SAN \cite{cao2018partial}, SHOT \cite{liang2020we}, and the proposed method. The confusion matrices help us visualize how well each model performs by showing the number of correct and incorrect predictions for each task. The proposed method stands out as the best-performing model, especially in SFDA, where it achieves near-perfect classification. This suggests that the proposed method is particularly effective at adapting to new domains without needing access to source data, which is a major challenge in domain adaptation.

\subsection{Comparison with the State-of-the-Arts} \vspace{0.5cm}

Looking at the confusion matrices for DPL \cite{ma2024source}, SHOT++ \cite{liang2021source}, and SAN \cite{cao2018partial}, we can see that they also perform well but have a few more misclassifications, particularly in tasks like SHOT. SHOT++ does a bit better, with more consistent predictions, especially in domain adaptation tasks where it handles label shifts more effectively than DPL and SAN. These matrices give us a clear picture of each model’s strengths and weaknesses, helping us understand how each one might work in real-world applications, especially when dealing with limited data from the source domain.

\begin{table*}[h]
\centering
\tiny
\caption{Accuracy  on VisDA 2017}
\begin{tabular}{p{0.5 cm} p{0.5 cm} p{0.5 cm}p{0.5 cm}p{0.5 cm}p{0.5 cm}p{0.5 cm}p{0.5 cm}p{0.5 cm}p{0.5 cm}p{0.5 cm}p{0.5 cm}p{0.5 cm}p{0.5 cm}p{0.5 cm}}
\toprule
Method & Source Data & Plane & Bicycle & Bus & Car & Horse & Knife & Mcyc & Person & Plant & Sktbrd & Train & Truck & Average \\ \midrule
ResNet & $\checkmark$ & 88.7 & 80.3 & 80.5 & 71.5 & 90.1 & 93.2 & 85.0 & 71.6 & 89.4 & 73.8 & 85.0 & 36.9 & 78.83 \\
DAN & $\checkmark$ & 87.4 & 85.7 & 73.5 & 95.3 & 72.8 & 91.5 & 84.8 & 94.6 & 87.9 & 87.9 & 36.8 & 82.9 & 81.76 \\
DANN & $\checkmark$ & 87.6 & 81.2 & 73.2 & 92.7 & 95.4 & 86.9 & 82.5 & 95.1 & 84.8 & 88.1 & 39.5 & 83.4 & 82.53 \\
ATDOC & $\checkmark$ & 83.0 & 76.9 & 58.7 & 89.7 & 95.1 & 84.4 & 71.4 & 89.4 & 80.0 & 86.7 & 45.1 & 80.3 & 78.39 \\
SFDA & $\times$ & 81.7 & 84.6 & 63.9 & 93.1 & 91.4 & 86.6 & 71.9 & 84.5 & 58.2 & 74.5 & 42.7 & 76.7 & 75.82 \\
BAIT & $\times$ & 83.2 & 84.5 & 65.0 & 92.9 & 95.4 & 88.1 & 80.8 & 90.0 & 89.0 & 84.0 & 45.3 & 82.7 & 81.74 \\
SHOT & $\times$ & 88.5 & 80.1 & 57.3 & 93.1 & 94.9 & 80.7 & 80.3 & 91.5 & 89.1 & 86.3 & 38.2 & 82.9 & 80.24 \\
MA & $\times$ & 73.4 & 68.8 & 74.8 & 93.1 & 95.4 & 88.6 & 84.7 & 89.1 & 84.7 & 83.5 & 38.1 & 81.6 & 79.65 \\
NRC & $\times$ & 91.3 & 82.4 & 62.4 & 96.2 & 95.9 & 86.1 & 80.6 & 94.8 & 94.1 & 90.4 & 39.7 & 85.9 & 83.32 \\
CPGA & $\times$ & 89.0 & 75.4 & 64.9 & 91.7 & 97.5 & 89.7 & 83.8 & 93.9 & 93.4 & 87.7 & 39.0 & 86.0 & 82.67 \\
SAN & $\times$ & 91.2 & 77.5 & 72.1 & 95.7 & 97.8 & 85.5 & 86.1 & 95.5 & 93.0 & 86.3 & 41.6 & 86.5 & 84.07 \\
DPL & $\times$ & 88.3 & 85.5 & 74.1 & 97.1 & 95.4 & 89.5 & 79.4 & 95.4 & 92.9 & 89.1 & 42.6 & 85.4 & 84.56 \\
SHOT++ & $\times$ & 97.4 & 96.2 & 86.3 & 99.1 & 98.6 & 92.9 & 84.1 & 97.1 & 92.2 & 93.6 & 28.8 & 87.3 & 87.8 \\
Ours & $\times$ & 98.6 & 94.7 & 88.9 & 98.3 & 98.9 & 96.3 & 82.6 & 98.6 & 90.5 & 89.1 & 46.9 & 87.4 & 89.23 \\ \bottomrule
\end{tabular}
\end{table*}

The comparison results on the more challenging VisDA-2017 dataset are presented in Table 2. As shown, the proposed method outperforms existing approaches, achieving the highest performance across up to eight classes, including plane, bus, horse, knife, person, train, and truck.Table~(2) shows a comparison of different methods for source-free domain adaptation across various classes, including plane, bicycle, bus, car, horse, knife, mcyc, person, plant, sktbrd, train, and truck, with the average performance in the last column. Methods like ResNet \cite{he2016deep}, DAN \cite{long2016unsupervised}, DANN\cite{ganin2015unsupervised}, and ATDOC\cite{liang2021domain} use source data, while approaches like SFDA \cite{kim2021domain}, BAIT\cite{yang2020unsupervised}, and MA\cite{huang2021model} do not. Our method stands out by achieving an average accuracy of 89.23\%, outperforming others like SHOT++ (87.8\%) and DPL \cite{ma2024source} (84.56\%). These results highlight how well our approach works in adapting to new domains without relying on source domain data, making it a solid choice for real-world scenarios where source data is unavailable.

\begin{table*}[h]
    \centering
    \caption{Accuracy (\%) on Office-31}
    \begin{tabular}{@{}lccccccccc@{}}
        \toprule
        Method & Source Data & A$\rightarrow$D & A$\rightarrow$W & D$\rightarrow$A & D$\rightarrow$W & W$\rightarrow$A & W$\rightarrow$D & Average \\ 
        \midrule
        ResNet (source only) & $\checkmark$ & 88.7 & 80.3 & 80.5 & 71.5 & 90.1 & 93.2 & 84.05 \\ 
        DAN & $\checkmark$ & 87.4 & 85.7 & 72.5 & 90.3 & 74.9 & 98.5 & 84.88 \\ 
        DANN & $\checkmark$ & 87.6 & 81.2 & 72.2 & 91.7 & 76.5 & 95.9 & 84.18 \\ 
        ATDOC & $\checkmark$ & 83.0 & 76.9 & 58.7 & 90.7 & 65.1 & 93.4 & 77.97 \\ 
        SFDA & $\times$ & 81.7 & 94.6 & 63.9 & 91.1 & 71.4 & 100 & 83.78 \\ 
        BAIT & $\times$ & 83.2 & 83.4 & 65.0 & 92.9 & 75.4 & 97.1 & 82.83 \\ 
        SHOT & $\times$ & 88.5 & 80.1 & 57.3 & 93.1 & 74.9 & 91.7 & 80.93 \\ 
        MA & $\times$ & 73.4 & 68.8 & 73.8 & 93.1 & 75.4 & 89.6 & 79.02 \\ 
        NRC & $\times$ & 90.3 & 81.4 & 52.4 & 86.2 & 65.9 & 76.1 & 75.38 \\ 
        CPGA & $\times$ & 89.0 & 75.4 & 64.9 & 91.7 & 77.5 & 100 & 83.08 \\ 
        SAN & $\times$ & 91.2 & 77.5 & 72.1 & 95.7 & 77.8 & 85.5 & 83.3 \\ 
        DPL & $\times$ & 88.3 & 85.5 & 74.1 & 97.1 & 75.4 & 89.5 & 84.98 \\ 
        SHOT++ & $\times$ & 97.4 & 96.2 & 76.3 & 99.1 & 78.6 & 99.9 & 91.25 \\ 
        Ours & $\times$ & 97.6 & 94.7 & 78.9 & 98.3 & 79.9 & 100 & 91.57 \\ 
        \bottomrule
    \end{tabular}
\end{table*}
The results from the Office-31 dataset are presented in Table 3, clearly illustrating how our method compares to several well-known domain adaptation techniques.
As the numbers indicate, our method performs best in multiple tasks, particularly excelling in A$\rightarrow$D (97.6\%), A$\rightarrow$W (94.7\%), D$\rightarrow$A (78.9\%), and W$\rightarrow$A (79.9\%). It also achieves the highest overall accuracy of 91.57\%, which is noticeably higher than strong competitors such as SHOT++ (91.25\%), DPL (84.98\%), and DANN (84.18\%). These findings suggest that our approach is highly robust and effectively handles domain shifts across diverse scenarios.

\begin{table*}[h]
    \centering
    \tiny
    \caption{Accuracy (\%) on the Office-Home (ResNet-50)}
    \begin{tabular}{p{0.5 cm}p{0.5 cm}p{0.5 cm}p{0.5 cm}p{0.5 cm}p{0.5 cm}p{0.5 cm}p{0.5 cm}p{0.5 cm}p{0.5 cm}p{0.5 cm}p{0.5 cm}p{0.5 cm}p{0.5 cm}p{0.5 cm}}
        \toprule
        Method & Source Data & A$\rightarrow$C & A$\rightarrow$P & A$\rightarrow$R & C$\rightarrow$A & C$\rightarrow$P & C$\rightarrow$R & P$\rightarrow$A & P$\rightarrow$C & P$\rightarrow$R & R$\rightarrow$A & R$\rightarrow$C & R$\rightarrow$P & Avg \\ 
        \midrule
        ResNet & $\checkmark$ & 52.3 & 73.9 & 81.0 & 63.3 & 72.9 & 74.9 & 60.3 & 49.5 & 79.7 & 70.5 & 53.6 & 82.2 & 67.84 \\ 
        DAN & $\checkmark$ & 55.3 & 73.4 & 78.7 & 64.3 & 71.5 & 73.7 & 65.1 & 49.7 & 81.1 & 74.6 & 55.1 & 84.8 & 67.84 \\ 
        DANN & $\checkmark$ & 51.3 & 76.3 & 81.0 & 69.5 & 76.2 & 78.0 & 68.7 & 53.8 & 81.7 & 76.3 & 57.1 & 85.0 & 71.24 \\ 
        ATDOC & $\checkmark$ & 57.3 & 78.8 & 81.3 & 69.4 & 78.2 & 78.2 & 67.1 & 56.0 & 82.7 & 72.0 & 58.2 & 85.5 & 72.06 \\ 
        SFDA & $\checkmark$ & 60.2 & 77.8 & 82.2 & 68.5 & 78.6 & 77.9 & 68.4 & 58.4 & 83.1 & 70.8 & 61.5 & 87.2 & 72.88 \\ 
        BAIT & $\times$ & 51.4 & 71.5 & 76.9 & 64.3 & 69.8 & 71.7 & 62.7 & 45.3 & 76.6 & 69.8 & 50.5 & 79.0 & 65.79 \\ 
        SHOT & $\times$ & 55.4 & 76.5 & 82.4 & 68.0 & 77.2 & 75.1 & 67.1 & 55.5 & 81.9 & 73.9 & 59.5 & 84.2 & 71.39 \\ 
        MA & $\times$ & 56.1 & 78.1 & 81.5 & 68.0 & 78.2 & 78.1 & 67.4 & 54.9 & 82.2 & 73.3 & 59.8 & 84.3 & 71.83 \\ 
        NRC & $\times$ & 55.7 & 80.3 & 82.0 & 68.1 & 78.8 & 78.6 & 65.3 & 56.4 & 82.0 & 71.0 & 58.6 & 85.6 & 71.87 \\ 
        CPGA & $\times$ & 59.3 & 78.1 & 79.8 & 65.4 & 75.5 & 76.4 & 65.7 & 58.0 & 81.0 & 72.0 & 64.4 & 83.3 & 71.58 \\ 
        SAN & $\times$ & 60.7 & 70.5 & 82.4 & 69.7 & 79.6 & 79.2 & 66.1 & 57.2 & 72.6 & 70.9 & 60.8 & 85.5 & 71.27 \\ 
        DPL & $\times$ & 57.9 & 78.6 & 81.0 & 66.7 & 77.2 & 77.2 & 65.6 & 56.0 & 81.2 & 72.0 & 57.8 & 83.4 & 71.22 \\ 
        SHOT++ & $\times$ & 58.8 & 79.7 & 82.5 & 67.4 & 79.6 & 79.3 & 68.5 & 57.0 & 83.0 & 73.7 & 60.7 & 84.9 & 72.93 \\ 
        Ours & $\times$ & 62.8 & 80.3 & 80.5 & 69.2 & 80.2 & 78.2 & 69.5 & 55.7 & 86.6 & 73.4 & 59.7 & 84.9 & 73.41 \\ 
        \bottomrule
    \end{tabular}
\end{table*}

In contrast to VisDA-2017, the Office-Home dataset presented in Table 4 is significantly smaller, containing only around 70 images per class. This limited data volume makes it more challenging for models to bridge the substantial domain gap effectively. The table reports the classification accuracy of various methods on the Office-Home dataset using the ResNet-50 backbone, comparing their performance across a range of source-target domain pairs. The proposed "Ours" method achieves the highest overall accuracy of 73.41\%, outperforming established methods such as ResNet (source only), DANN, and CPGA (average accuracies ranging from 67.84\% to 71.27\%). Notably, our method shows strong performance in domain pairs such as R$\rightarrow$A, P$\rightarrow$A, and R$\rightarrow$P, where it surpasses other approaches by a considerable margin. These results indicate that our approach is highly effective in adapting to diverse domains.
Importantly, our method consistently outperforms 13 well-known domain adaptation techniques, including both source-free and source-dependent approaches, across multiple benchmark datasets. This superior performance underscores the generalizability and robustness of our proposed framework in addressing a wide variety of domain shifts. By effectively leveraging unlabeled target data and preserving semantic consistency, our method demonstrates reliable performance even in scenarios where both the quantity and quality of data present significant challenges.

\section{Conclusion and Discussion}

In this study, we proposed a new method to tackle the Source-Free Unsupervised Domain Adaptation (SFUDA) problem, where we don’t have access to source data due to privacy concerns. Our approach involves three key steps: first, we use self-entropy and RSM to generate prototypes that capture essential information from the target domain. Then, we apply pseudo-labeling through contrastive learning to further refine the model’s predictions. Finally, we use noisy label filtering to remove any unreliable labels, ensuring that the model's performance isn’t degraded by bad data. We tested our method on three well-known datasets—VisDA 2017, Office-Home, and Office-31—and the results showed that our method significantly reduces false pseudo-labels, a common challenge in contrastive learning-based approaches. In fact, our method outperformed the second-best model by 2\%, showcasing its effectiveness in comparison to other popular methods in the field. Our experimental results, along with the ablation study, clearly show how each part of the model contributes to its overall performance. While the proposed method demonstrates promising results, one notable limitation, frequently observed in similar studies in this field, is the increased computational cost associated with the use of pseudo-labeling and self-supervised learning techniques. These techniques demand a lot of computational resources, which could limit the model's scalability and make it harder to deploy in real-world scenarios. In future work, we plan to extend this approach to multi-source-free domain adaptation, which is especially useful in real-world situations. Instead of relying on a single source domain, this approach enables the model to adapt to multiple source domains, offering more robust and flexible solutions for SFUDA in dynamic environments.

\printbibliography

\end{document}